# Curved Space Optimization: A Random Search based on General Relativity Theory

Fereydoun Farrahi Moghaddam; Reza Farrahi Moghaddam, *Member, IEEE*; and Mohamed Cheriet, *Senior Member, IEEE.*

*Abstract*—Designing a fast and efficient optimization method with local optima avoidance capability on a variety of optimization problems is still an open problem for many researchers. In this work, the concept of a new global optimization method with an open implementation area is introduced as a Curved Space Optimization (CSO) method, which is a simple probabilistic optimization method enhanced by concepts of general relativity theory. To address global optimization challenges such as performance and convergence, this new method is designed based on transformation of a random search space into a new search space based on concepts of space-time curvature in general relativity theory. In order to evaluate the performance of our proposed method, an implementation of CSO is deployed and its results are compared on benchmark functions with state-of-the-art optimization methods. The results show that the performance of CSO is promising on unimodal and multimodal benchmark functions with different search space dimension sizes.

## I. Introduction

An efficient Global Optimization Method (GOM) should be able to avoid local optima and reach the global optimum with a low cost in terms of CPU cycles. It should also be easy to implement, and robust for different optimization problems.

However, there is always a tradeoff between avoiding local optima and global optimum convergence in global optimization methods (the exploration-exploitation tradeoff). When methods focus on exploitation, the chance of becoming trapped in local optima is high, and when they focus on exploration, the methods efficiency is low and it is slow to converge to the optimal value. Usually, GOMs start with exploration and then gradually move to exploitation. Early exploration will try to find, as much as possible, areas in the search space with higher potential for global optimum existence, and then exploitation will guarantee convergence.

For example, in the Simulated Annealing (SA) method, the early high temperature will raise the exploration capability of the method, and the later cooling mechanism will guarantee convergence. If a method does not spend enough time on exploration, it will fall into local optima, and, if it spends too much time on exploration, its efficiency will decrease.

Fereydoun Farrahi Moghaddam, R. Farrahi Moghaddam and M. Cheriet are with the Synchromedia Laboratory for Multimedia Communication in Telepresence, École de Technologie Supérieure, Montreal, (QC), Canada H3C 1K3; Tel.: +1-514-396-8972; Fax: +1-514-396-8595; Emails: ffarrahi@synchromedia.ca, reza.farrahi@synchromedia.ca, mohamed.cheriet@etsmtl.ca

There are other optimization methods, like Genetic Algorithms, which are different from SA as they work with a population of points from the search space. To avoid early convergence, they take advantage of mutation operators, which add the exploration capability to their convergent nature. In other famous GOMs, such as Particle Swarm Optimization (PSO), Ant Colony Optimization (ACO), Evolutionary Programming (EP), and Tabu Search (TS), exploration is provided by random parameters ($r_p$ and $r_g$), the pheromone evaporation process, the mutation operator, and space memory respectively. For most of the population-based GOMs, the initial exploration feature is provided by an initial distributed population across the search space. For the Random Search there is no convergence mechanism, and the Hill Climber method is a one hundred percent convergence-based method.

All GOMs have some parameters, the proper adjustment of which is critical for providing a balance between exploration and exploitation in that particular method. For example, when the mutation rate is high in the GA, the algorithm will avoid early convergence better than a lower mutation rate, but the performance of the algorithm will decrease accordingly. The same argument is true for ACO, when the pheromone evaporation rate is high.

To address the challenges of the global optimization problem, such as performance and convergence, a new GOM is introduced in this research based on a combination of a simple search method and a physical phenomenon theory. The simple search method, which is good at exploration by its very nature, will be helped by the physical phenomenon theory to improve its exploitation capabilities, in order to achieve high performance and global optimum convergence goals. To find a balance for the new algorithm between exploration and exploitation, self adaptive mechanisms are used in the selection of the parameters of the new method.

The rest of article is organized as follows. First, related work on global optimization is discussed in section II-A and the concept of general relativity theory, which is used in the CSO method, is explained in section II-B. Then, the concept of CSO is introduced in section III and its infinite number of implementation strategies are discussed, followed by the introduction of a simple and basic CSO implementation in section IV, which is used to evaluate the new method. Finally, in section V, the CSO method is tested on state-of-the-art benchmark functions, which are described in section V-A, and its results are compared with those of other GOM in section V-B.

## II. Related Work

*A. Global Optimization Methods*

GOMs are used to find the optimal value of a function, regardless of any initial preference. These methods can be categorized in two main classes: deterministic and probabilistic [1]. The probabilistic methods are mostly categorized as: i) stochastic; and ii) heuristic. There are many methods for finding the optimal point of functions, but they differ in their success rate and in their performance on different types of problems. No one method acts perfectly on all types of functions. Acting perfectly means avoiding local optima and promising a global optimum point, while keeping the processing time and number of evaluations small. For example, Random Search [2], which is the most basic stochastic technique, can guarantee the global optimum, but it is very time consuming. The Hill Climbing technique, which is a basic deterministic method, is very quick, but cannot guarantee the global optimum. These two techniques work on a single point, while most probabilistic methods, such as the Genetic Algorithm and the Ant Colony, have a population of points providing different solutions to the problem. Below, a brief discussion of some of the best known GOMs is provided.

*1) Particle Swarm Optimization (PSO)::* In 1995, Kennedy and Eberhart introduced PSO, which is inspired by the swarming behavior displayed by a flock of birds or a school of fish, or even human social behavior under the influence of other individuals [3], [4], [5], [6]. Practically speaking, PSO simulates a swarm of particles moving in a multidimensional search space of real values. Every particle, each representing a possible solution, has a position vector and a velocity vector. Moreover, each particle stores a small amount of information about its own best position gathered so far. The whole swarm also stores the global best position, which is available to all particles. The velocity and position of each particle is updated based on a governing equation that depends on the best position of the particle, as well as the global best position and a few random coefficients [5].

The main difference between PSO and evolutionary computation optimizers is that PSO does not have an explicit selection function [7] that reallocates the search resources to new individuals that will potentially perform well. Instead, each particle in the PSO behaves based on its personal best position. It can be said that the PSO implicitly has twice the population. Also, the way parent information is handled is different in the PSO; it is contained and manipulated within each particle, in contrast to evolutionary optimizations where this information is shared [7]. However, the PSOs highly directional governing equations can reduce its performance on some problems, such as the Griewank function [7].

*2) Simulated Annealing (SA)::* SA is one of the heuristic algorithms, and is based on the metallurgic annealing process [8], [9], [10]. This algorithm simulates a collection of metal atoms in equilibrium at a certain temperature, and uses iterative local searching with an individual acceptance criterion. It starts with an initial state, and, in each iteration, the change in energy after a random change in this state is calculated. Then, if the energy change is negative, the new state is carried into the next iteration. However, even if the new state has a higher energy, it is possible for it to be carried forward into the next iteration. This is performed by a probability determined by temperature, the energy change, and Boltzmanns constant. The case of positive energy change can be seen as an uphill move, and is analogous to a higher energy molecule knocking loose a molecule trapped in a state of excess energy. This probability enables the algorithm to avoid being trapped in the local minima. A key factor in the success of SA is its annealing temperature, which is a variable that decreases in time. This decrease in temperature means a decrease in the probability of uphill moves, which helps the algorithm to converge. Determining the cooling (annealing) scheme of SA is crucial to the success of the algorithm. With intelligent cooling, the algorithm can escape local minima and reach the global minimum. The algorithm details, which should be set for each problem, including initialization, neighbor solutions, temperature and cooling scheme initialization, and a stopping criterion. It has be shown that SA can be used in both adaptive and continuous optimization [11].

*3) Ant Colony Optimization (ACO)::* In the early 1990s, inspired by the research done on real ants [12], M. Dorigo and his colleagues introduced the Ant System [13]. Over time, many other similar algorithms have been proposed, among them the MAX-MIN Ant System [14] and the Ant Colony System [15]. All these algorithms, for example [16], [17], [18], are categorized as ACO metaheuristics, which provide solutions for hard combinatorial optimization problems, or any problem that can be reduced to finding an optimal path in a (directed) graph [19]. In their search for food, ants start to randomly navigate in the area around their nest. Similar to a random walk, if an ant cannot find food, its distance to the nest will increase over time. When an ant finds a source of food, it returns as much of it as possible to the nest, and at the same time it colors its track with the pheromone coded with that food. If other ants come across to its track, they can decide whether or not to switch to that track. This will increase the amount of the pheromone on the track, and eventually a majority of the ants will participate in the retrieval of food from that source. However, because of the probabilistic nature of ant decision making, there are always some ants that search for other food sources. Also, the pheromone has a limited half-life, and so the track will disappear when the food supply is depleted at that source. Although the ACO is an optimizer for graph-based problems, many real vector optimization problems can be rewritten as a graph problem, and can therefore be solved by the ACO [20]. However, there are many parameters and variations in ACO algorithms, and so using it requires expertise, which varies from one application to another, on these parameters and variations.

*4) Evolutionary Programming (EP)::* In 1964, L. J. Fogel introduced evolutionary programming (EP) [21], which came to be used to develop artificial intelligence based on finite state machines as predictors for data streams [22], [23], [19]. A key feature of EP is that it assumes that a solution candidate is a species without any crossover with other species. In this way, mutation is the only evolutionary mechanism. At each iteration, a parent generates just one offspring. This can be formulated as the ($\mu$+$\mu$) evolution strategy. In this latter strategy, from a set of $\mu$ parents, $\mu$ offspring are created, and then, from the joint set of parents and offspring, a set of $\mu$ new parents is kept based on their fitness. Another way to look at EP is to assume that it considers a fixed organization and structure of a problem, and allows only parameter values to change and evolve. EP has been used in many learning and optimization applications. For example, in [24], EP was used for the blind equalization of signals transmitted over a distorting channel after the problem had been converted to the optimization of a multimodal, multivariable cost function.

*5) Tabu Search (TS):*: This approach avoids the problem of becoming trapped in a local minimum by generalizing and extending Hill Climbing with the concept of sacred tabu[1] points [25], [26], [27], [28], [29], [30]. In TS, the solution candidates already visited will not be visited again, because they have been labeled as tabu. In some variation of TS, a neighborhood around the visited candidate is declared as a tabu area. To keep track of all the candidates visited, TS maintains a list of them. However, because of limited resources, the list is kept as short as possible, and is updated based on the fitness of new candidates. At the same time, aspiration criteria are considered in TS, in order to control the impact of tabu-labeled candidates on the performance of the method. Without these criteria, it is highly probable that the candidates, which are supposed to prevent the method from becoming trapped in a local minimum, could prevent the method from approaching the region in the search space that contains the global minimum. In its simplest form, an aspiration criterion temporarily replaces the tabu list with an empty list to allow a candidate to be selected free from the tabu restriction. This candidate survives if its fitness is better than that of the current best known candidate. Several variations of TS have been developed, mainly by hybridization with other search methods [31], and it has been successfully used as a global optimizer in many applications [32], [33].

*B. Space-Time Curvature in General Relativity Theory*

In this subsection, we briefly present how our proposed method, which is based of curved spaces, has been inspired by the space-time curvature (or space curvature, as it is generally known) of general relativity theory. Einsteins proposal of space curvature in his general relativity theory was one of the most radical theoretical steps ever taken, and it exposed the myth of absolute time [34]. The other major novelty of general relativity theory was its ability to relate fundamental concepts, such as gravitational mass and inertial mass, which had previously been thought to be completely independent [35], [36]. In short, the presence of a mass influences the space around it, and converts the flat space-time of a vacuum into curved space-time. Curved space-time cannot then be considered independent from mass and energy. In general relativity theory, the space-time dynamic is modeled by the *material* energy tensor $T_{ik}$, the Riemann curvature tensor $R_{ik}$,

[1]A prohibition excluding an object being used or touched because of its inviolable or sacred nature.

and the metric tensor $g_{ik}$ [37], [34], as follows:

$$R_{ik} - \frac{1}{2}g_{ik}R = -T_{ik}, \ i,k = 1,2,\cdots,4 \quad (1)$$

$T_{ik}$ is called the material energy tensor to emphasize the fact that it does not exclusively depend on $g_{ik}$. Usually, the term $T_{i=0,k=0}$ is considered as the mass or energy density. Obviously, Equation (1) shows a direct relation between $R_{ik}$ and $T_{00}$, which are the space-time curvature and the mass density respectively. At the same time, the presence of a particle is also related to $T_{00}$ [38].

## III. CURVED SPACE OPTIMIZATION

As mentioned in section II-A, Random Search is a very slow method, but with it there is no risk of falling into a local minimum. In the proposed Curved Space Optimization (CSO), the curvature of space inspired by general relativity theory is used to improve the efficiency of a simple random search, and convert it to a very robust optimization tool. Indeed, the correlation between the space-time curvature and the presence of a particle in physics, which was described in section II-B, inspired us to develop a curved space-based optimization method in which the minima play the role of particles. If we consider the search space of an optimization problem as a generalized space-time space similar to that of general relativity theory, we can translate the presence of a particle into the presence of a local minimum point. Therefore, the relation between the space-time curvature and the presence of a particle suggests that there is a greater chance of finding a local minimum point if we look at those regions of the search space that have higher curvature. In this way, the search for the global minimum point, which can be seen as crawling among all the possible local minima and digging into the search space around those possible candidates, can be guided by the curvature of the search space.

As mentioned before, CSO is based on the Random Search method, and it will behave like RS to some extent. In RS, the search space is searched uniformly to find the optimum value. Unlike other methods, such as GA or SA, there is no memory in RS, and so selecting the new points bears no relation to the areas already searched. As shown in Figure 1, in a two-dimensional space, the search spots are uniformly speared on the surface of the search space. This explains the slowness of RS. It also explains why RS will never fall into a local minimum.

In CSO, the search space will be bent according to the value of a spot already searched, in just the same way as space-time is bent in the presence of a mass in general relativity theory. In this theory, the curvature of space-time grows when the mass is grows larger. The same thing happens in CSO: the search space will be bent more if the fitness value of a searched spot is higher, and will be bent less if the fitness value is smaller. In Figure 2, the same two-dimensional search space of the previous example is bent under two spots with different fitness values.

While the search is going on, more spots are searched, and there will be more curvature on the search space. This curved space can be used as a guide for RS to select new spots. The

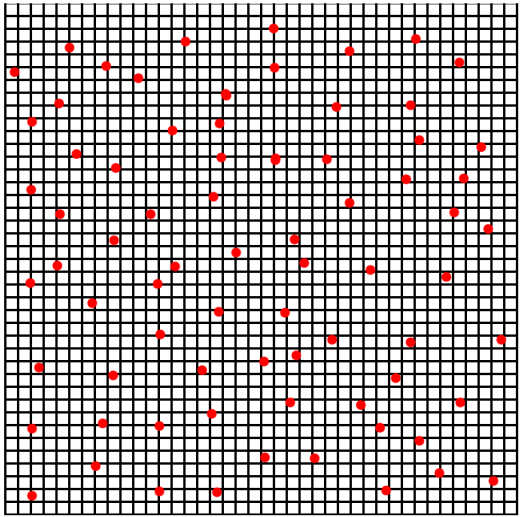

Fig. 1. The uniform distribution of searched points in the Random Search method

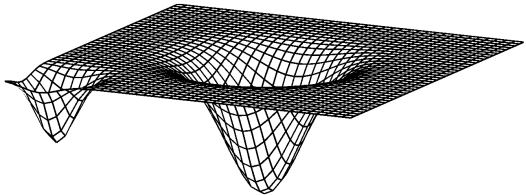

Fig. 2. A curved search space

areas with higher space curvature show higher probability of finding spots with better fitness. In Figure 3, the sample search space is shown after several curvature iterations. As shown, the curved search space will reveal the shape of a test function after enough search iterations. In Figure 3, the test function is $z = x^2 + y^2$

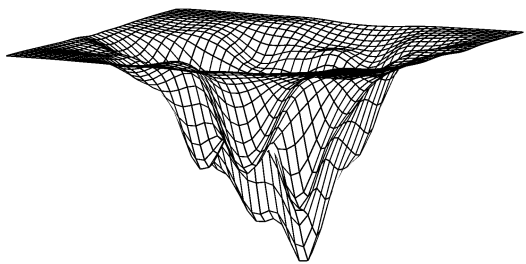

Fig. 3. Curved search space after a number of iterations

If RS is performed in the new search space, it will help to find finer points with respect to the shape of the curved space. Considering the relation between the presence of a particle and the curvature of space-time, the search process can be seen as a curvature analysis of space-time. Not only does the curvature serve as a guide towards possible candidates, it also allows us to analyze the quality of minimum points in the proximity. Because of the high complexity and cost of direct analysis of the curved search space, we propose an indirect way to map the curved search space to a new flat search space in which traditional random and heuristic search and optimization methods can easily be used. The mapping between the curved and the flat spaces automatically enforces more crawling around the high quality local minima, which enables the search and optimization algorithms to converge to the global minimum point without becoming trapped in the local minima. Therefore, in order to perform RS in a new curved search space, that space needs to be flattened. Once new points have been selected, they need to be transformed back to the non flattened search space. The process of selecting new points in a new space is depicted in Figures 4, 5, and 6.

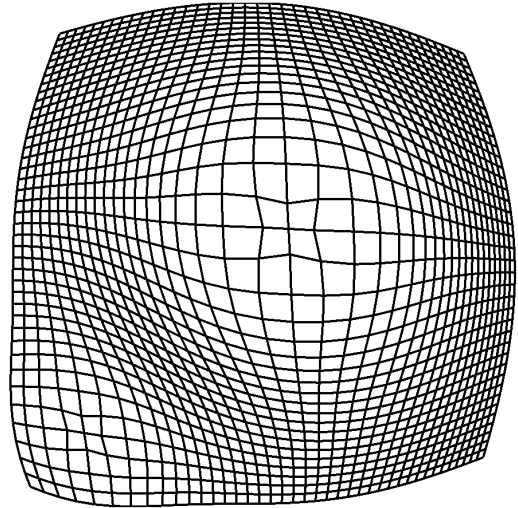

Fig. 4. The flattened curved search space.

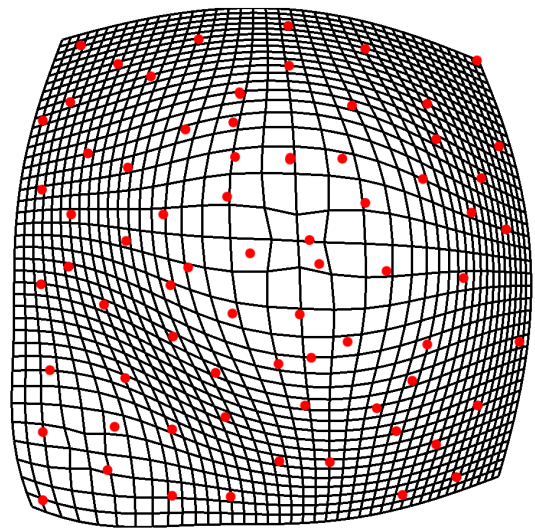

Fig. 5. A set of uniformly chosen points in the flattened search space.

As can be seen in Figures 5 and 6, the selected points in the normal search space are denser in areas with higher curvature and less dense in other areas. RS still preserves its good features, but is more guided and focused on areas with higher curvature, where there is a higher probability of finding points with better fitness values. This is still RS, but guided by curvatures of the search space. A higher curvature of space at a spot means higher density of new random search points

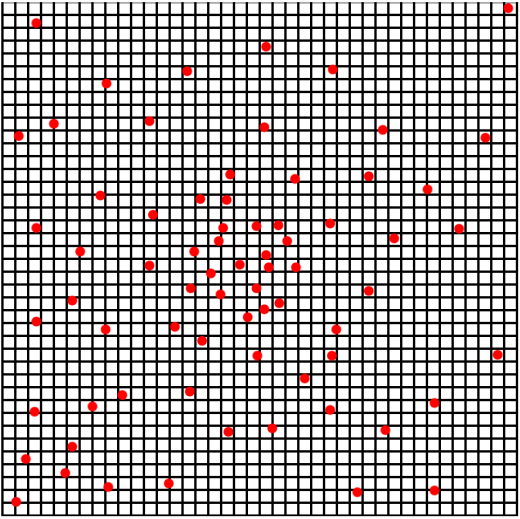

Fig. 6. New search points transformed from the flattened search space into the ordinary search space.

landing in that area.

Although the implementation of this method is complex, the concept of CSO can be represented as a simple pseudocode, as shown in Algorithm 1.

**Algorithm 1:** CSO algorithm in pseudocode:
1 **repeat**
2    create the curved-search-space based on previously selected points;
3    flatten the curved-search-space;
4    choose an uniform random point in flattened-curved-search-space;
5    transform the selected point to the ordinary search space;
6    evaluate the fitness of selected point;
7 **until** *terminating condition*;

Various methods can be used to make the curvature in the space, to flatten the curved space, and to transform the points between spaces. Choosing the appropriate method can have a big impact on the performance and speed of the algorithm. The depth of curvature is also very important. More research is needed to study the impact of different methods on the behavior of CSO, which is beyond the scope of this work. In section IV, a simple, axis-independent flattening method is used to implement the general features of the CSO. Based on the fundamental nature of CSO, which is inspired by RS, it is expected that this method will perform well in terms of avoiding the local minima. It is also expected that CSO will reach the global optimal point very quickly under the guidance of space curvatures. Finally, it is expected that CSO will be easy to implement and fast to run, based on the simple pseudocode of the algorithm. But, as mentioned before, implementing the space curvature, flattening the search space, and transforming the curved space can all have an impact on the performance and speed of the algorithm. The performance, speed, and implementation of CSO are discussed in section V, and its results compared with the other existing GOMs in the following sections.

## IV. CSO Implementation

To implement CSO, the type of curvature of the search space first needs to be defined. As mentioned in section III, many types of curvature can be used, some of which are shown in Figure 7.

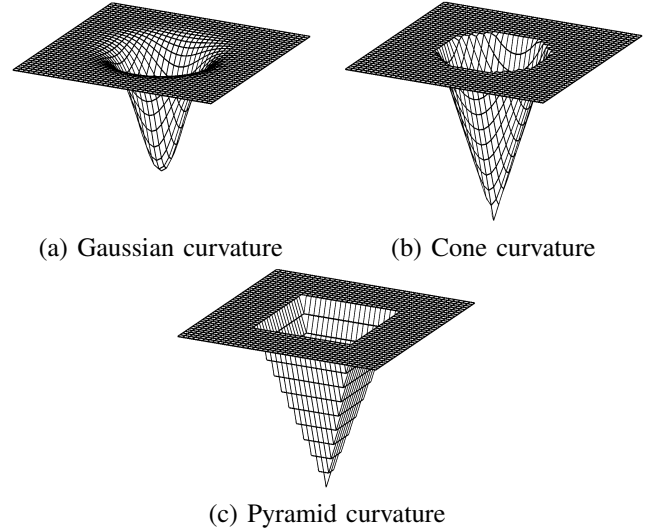

(a) Gaussian curvature    (b) Cone curvature

(c) Pyramid curvature

Fig. 7. Various possible types of curvature.

Regardless of the type of curvature used, the algorithm should define the depth and radius of the curvature in the Cone and Cylinder case, and $\sigma$ in the Gaussian case. In this study, a simple Ant Colony-based algorithm is used to choose either the size of the radius or $\sigma$. The depth of curvature is calculated based on the relative fitness values of the point. These mechanisms are described in sections IV-A and IV-B.

In this research, an Axis-Independent Gaussian Curvature (AIGC) is used for the space curvature, as depicted in Figure 8.

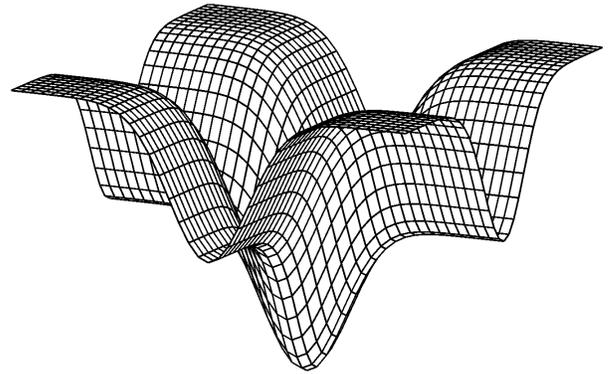

Fig. 8. The Axis-Independent Gaussian Curvature concept.

There are two advantages of using the AIGC. First, the flattening process is easy and quick. Second, not only does the space have its largest curvature near the point, but it is also slightly curved along the constant variable lines. These extra curvatures will help the algorithm find the optimal point faster if the variables are independent. However, even if the variables are not independent, which is the usual case, the largest curvature will help find the optimum point. The flattened version of the AIGC is depicted in Figure 9.

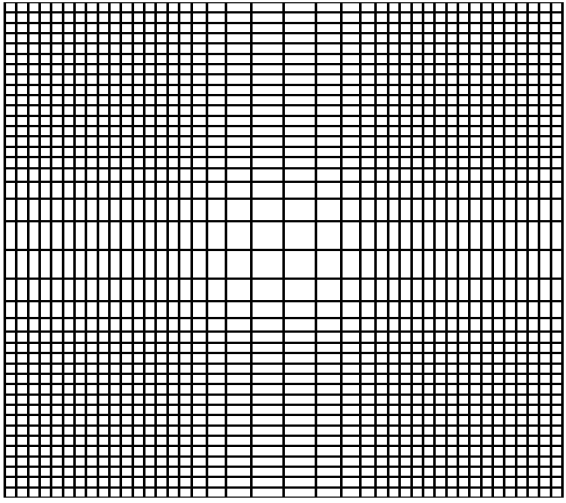

Fig. 9. Flattened Axis-Independent Gaussian Curvature

In a normal RS, the variables for new points are selected independently. The same is true for the AIGC implementation. When there is one axis, and after a few iterations, the shape of the axis curvature is a graph that constitutes a Gaussian mixture, as shown in Figure 10.

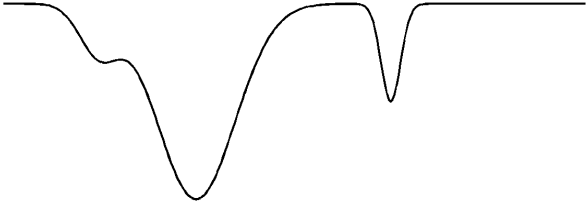

Fig. 10. One-axis curvature after a few iterations.

To choose a new point in the flattened space, it is sufficient to calculate the length of the curved axis and select a random point of that length, and then calculate the selected point in a normal axis using a reverse function of the length.

The equation of the curvature of each axis can be formulated as follows:

$$C_{x_i}(x) = -\sum_{p \in \mathbb{P}} \frac{a_p}{\sqrt{2\pi\sigma_p^2}} e^{\frac{-(x-\mu_p)^2}{2\sigma_p^2}} \quad (2)$$

where $\mathbb{P}$ represents the set of points where curvature accrued. $\mu_p$ shows the value of the variable $x_i$ at point $p$, $\sigma_p$ shows the sharpness of curvature, and $a_p$ shows the depth of curvature. Below, Equation (3) represents the length of curved axes:

$$\begin{aligned} l_{C_{x_i}}(x) &= \int_{x_{i_{min}}}^{x} \sqrt{1 + C'^2_{x_i}(y)} dy \\ &= \int_{x_{i_{min}}}^{x} \sqrt{1 + \left(\sum_{p \in \mathbb{P}} a_p \frac{(x-\mu_p)}{\sqrt{2\pi}\sigma_p^3} e^{\frac{-(x-\mu_p)^2}{2\sigma_p^2}}\right)^2} dy \end{aligned} \quad (3)$$

The total length of the $i^{th}$ curved axes is $l_{C_{x_i}}(x_{max})$. When a random number between 0 and $l_{C_{x_i}}(x_{i_{max}})$ is chosen, such as $x_c$, the reverse function of $l_{C_{x_i}}(x)$ can be used in order to calculate the value of a selected point in the normal axis, as follows:

$$x = l^{inv}_{C_{x_i}}(x_c) \quad (4)$$

As it is impossible to provide an exact mathematical formulation for $l_{C_{x_i}}(x)$ and $l^{inv}_{C_{x_i}}(x_c)$, numerical methods are used to calculate the values of these functions.

As shown in Equation (2), there are two types of parameter that control the curvature function $C_{x_i}(x)$, which are $a_p$ and $\sigma_p$. Very deep and sharp curvatures could pose a problem for exploration of the algorithm, while shallow curvatures could affect the performance of the algorithm. In the following sections, simple mechanisms for selecting efficient $a_p$ for each curvature and $\sigma_p$ for each iteration are explained.

### A. Mechanism for selecting depth of curvature

To select a depth value for a curvature, which is the $a_p$ parameter in Equation 3, the fitness value of the point will be used along all the other fitness values. Because the optimum value of the benchmarking function is not known to the algorithm, the fitness value of each point should be evaluated relative to all the fitness values of the other points. First, all the fitness values will be reevaluated by the worst fitness value among the recorded values. To minimize the benchmarking function, the following formula will be used:

$D_i = \mathcal{F}_{max} - \mathcal{F}_i, \; \forall i \in \{1, 2, \cdots, n\}$

Second, all the depth values will be normalized, as follows:
$D_i = D_i/D_{max}, \; \forall i \in \{1, 2, \cdots, n\}$

Finally, the depth value will be transformed with a focus function. The purpose of the focus function is to make the points with better fitness much more valuable than the points with worse fitness.

$D_i := \text{focus}(D_i), \forall i \in \{1, 2, \cdots, n\}$

This focus function can be as simple as a line or a power curve, as shown in Figure 11.

Using a sharper focus function, such as $f(x) = x^{10}$, the convergence process can be speeded up. However, smoother focus functions make the exploration easier for the algorithm, which is very important in multimodal cost functions. In this article, $f(x) = x^{10}$ is used as a focus function for all results, however further research is needed on this parameter to understand the behavior of the algorithm on different focus functions, and on the possibility of using adaptive focus functions in CSO.

### B. Ant Colony-based Width Selection Mechanism

As mentioned in section IV, one of important features of the CSO algorithm is the radius selection of the curvatures. A small radius of curvature will help convergence, and a large radius of curvature will help the algorithm explore more optimum valleys. In this implementation, the width of the curvature, which is equal to $\sigma_p$ in Equation (3), will be selected among 40 predefined widths (to provide a resolution of $10^{-20}$; for finer resolutions, a higher number should be considered) in a normalized search space. These predefined widths range from 1 to $10^{-19.5}$, as follows:

$$\sigma_i = 10^{\frac{-(i-1)}{2}}, i = 1, 2, ..., 40 \qquad (5)$$

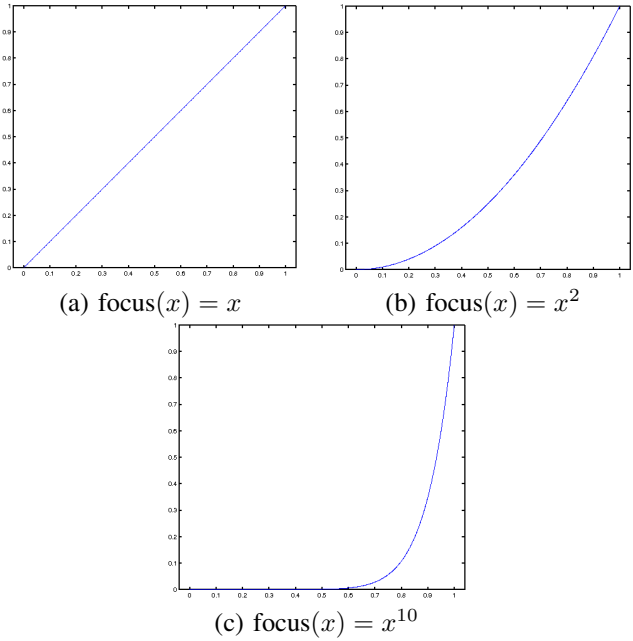

Fig. 11. Different focus functions.

In each iteration of the CSO algorithm, a different width will be chosen from these predefined widths. The process of width selection is based on an adaptive Ant Colony mechanism. In the beginning, the width will be selected randomly from all available widths. If the selected width leads to a successful iteration, an amount of pheromone will be added to the coresponding width. For later iterations, the probability of width selection will be based on the amount of pheromone collected for each predefined width. The success of width selection is measured based on comparing the new point fitness value with the fitness values of previously stored points. The amount of pheromone is positively correlated to the success rate. For instance, the amount of pheromone is the difference between the fitness value of the new point and the worthworst stored fitness value. The pheromone collected for each predefined width evaporates at a low rate, which is different for each width. The reasons why evaporation rates differ are given in the next section.

With this mechanism, CSO will choose the best curvature width automatically. If the success rate on one of the curvature widths is better, the amount of pheromone collected will be greater for that curvature rate, and, as a result, this curvature width will be selected more often. This will help the algorithm focus on curvature widths that are more efficient. If that curvature width is no longer efficient, the success rate will be lower and the amount of pheromone for that curvature rate will evaporate, and so the CSO algorithm will switch to another efficient curvature width.

There are some points to consider with respect to curvature rate selection if early Ant Colony convergence is to be avoided. These are described in the next section.

### C. Other Points to Consider in CSO Implementation

The following points should be considered in CSO implementation in order to improve its performance:

*a) Curvature Width Selection Adjustment:* As mentioned in section IV-B, a curvature width will be selected from predefined widths according to the amount of pheromone collected for each width. In order to avoid early convergence in the Ant Colony algorithm, a randomly selected number (-1, 0, or 1) will be added to the selected width. This is a small mutation for ant colony algorithm to avoid becoming trapped in one width.

*b) Random Axis Curvature:* In order to boost the effect of space curvatures, a random set of axes will be selected in each iteration to be curved, instead of curving all the axes. This will enhance the performance of the algorithm.

*c) Larger Widths vs. Smaller Widths:* In CSO implementation, larger widths are boosted more than smaller ones in three ways. (This is simply to avoid early convergence of the algorithm.) First, naturally larger widths cause larger improvements, which means more pheromones. Second, to boost the larger widths even more, different evaporation rates are considered for different widths. Larger widths have smaller evaporation rates, and smaller widths have larger evaporation rates. Third, at the start when there is no pheromone present, a boosted bias function will be used, instead of a flat random selection of widths, to select the width that will choose slightly larger widths more often than smaller widths.

## V. EXPERIMENTAL RESULTS

In this section, some of the standard functions used for benchmarking GOMs in the literature are selected, in order to evaluate the performance of the CSO algorithm. Then, a simple and basic implementation of CSO, which was described in a previous section, is tested on the selected benchmark functions. Finally, the results of a few other methods which have been tested on these benchmark functions in the literature are listed and compared with the results of the proposed algorithm.

### A. Benchmark Functions

In this section, a variety benchmarking functions are selected from the literature to cover all the functions, and, for each function, a brief description is provided. Also, if the optimum point of the function is known, it is presented with the fitness value of the function at that point.

*1) Sphere function:* The sphere function is the simplest benchmark function. The optimal value for the sphere function is at $x_i = 0, i = 1, 2, 3, \cdots$, and the optimum value is 0.

$$f_{sph}(x) = \sum_{i=1}^{n} x_i^2 \qquad (6)$$

*2) Axis parallel hyper-ellipsoid function:* The optimal value for the parallel axis hyper-ellipsoid function is at $x_i = 0, i = 1, 2, 3, \cdots$, and the optimum value is 0. The search space is usually restricted to the hypercube $-5.12 \leq x_i \leq 5.12$.

$$f_{aph}(x) = \sum_{i=1}^{n}(ix_i^2) \quad (7)$$

*3) Rotated hyper-ellipsoid function:* The optimal value for rotated hyper-ellipsoid function is at $x_i = 0, i = 1, 2, 3, \cdots$, and the optimum value is 0. The search space is usually restricted to the hypercube $-65.536 \leq x_i \leq 65.536$.

$$f_{rhe}(x) = \sum_{i=1}^{n}\sum_{j=1}^{i}(x_j^2) \quad (8)$$

*4) Rosenbrock's function:* The optimum value for Rosenbrock's function is at 0. The search space is usually restricted to the hypercube $-2.048 \leq x_i \leq 2.048$.

$$f_{ros}(x) = \sum_{i=1}^{n-1}[100(x_{(i+1)} - x_i^2)^2 + (1-x_i)^2] \quad (9)$$

*5) Rastrigin's function:* The optimal value for Rastrigin's function is at $x_i = 0, i = 1, 2, 3, ...$ and the optimum value is 0. The search space is usually restricted to the hypercube $-5.12 \leq x_i \leq 5.12$.

$$f_{ras}(x) = 10n + \sum_{i=1}^{n}[x_i^2 - 10Cos(2\pi x_i)] \quad (10)$$

*6) Schwefel's function:* Schwefel's function is a multimodal function which its optimum point is not at $x_i = 0, i = 1, 2, 3, ....$ The optimal value for Schwefel's function is at $x_i = 420.9687$ and the optimum value is $-418.9829n$. The search space is usually restricted to the hypercube $-500 \leq x_i \leq 500$.

$$f_{sch}(x) = \sum_{i=1}^{n}[-x_i Sin(\sqrt{|x_i|})] \quad (11)$$

*7) Griewangk's function:* The optimal value for Griewangk's function is at $x_i = 0, i = 1, 2, 3, ...$ and the optimum value is 0. The search space is usually restricted to the hypercube $-600 \leq x_i \leq 600$.

$$f_{gri}(x) = \frac{1}{4000}\sum_{i=1}^{n}x_i^2 - \prod_{i=1}^{n}Cos(\frac{x_i}{\sqrt{i}}) + 1 \quad (12)$$

*8) Sum of different power function:* The optimal value for Sum Of Different Power function is at $x_i = 0, i = 1, 2, 3, ...$ and the optimum value is 0. The search space is usually restricted to the hypercube $-1 \leq x_i \leq 1$.

$$f_{sdp}(x) = \sum_{i=1}^{n}|x_i|^{i+1} \quad (13)$$

*9) Ackley's function:* The optimal value for Ackley's function is at $x_i = 0, i = 1, 2, 3, ...$ and the optimum value is 0. The search space is usually restricted to the hypercube $-32.768 \leq x_i \leq 32.768$.

$$f_{ack}(x) = -ae^{-b\sqrt{\frac{1}{n}\sum_{i=1}^{n}x_i^2}} - e^{\frac{1}{n}\sum_{i=1}^{n}Cos(cx_i)} + a + e^1 \quad (14)$$

where $a = 20$, $b = 0.2$, and $c = 2\pi$

*10) Michalewicz's function:* For $n = 5$, the optimum value for Michalewicz's function is $-4.687$. For $n = 10$, the optimum value for Michalewicz's function is $-9.66$. The search space is usually restricted to the hypercube $0 \leq x_i \leq \pi$.

$$f_{mic}(x) = -\sum_{i=1}^{n}Sin(x_i)[Sin(\frac{ix_i^2}{\pi})]^{2m} \quad (15)$$

where $m = 10$.

*11) 2n-minima function:* The optimal value for Ackley's function is at $-78.3323$. The search space is usually restricted to the hypercube $-5 \leq x_i \leq 5$.

$$f_{2nm}(x) = \frac{1}{n}\sum_{i=1}^{n}(x_i^4 - 16x_i^2 + 5x_i) \quad (16)$$

*12) Branins's function:* The optimal value for Branins's function is at $(x_1, x_2) = (\pi, 12.275), (\pi, 2.275), (9.42478, 2.475)$ and the optimum value is $0.397887$. The search space is usually restricted to $-5 \leq x_1 \leq 10$, $0 \leq x_2 \leq 15$.

$$f_{bra}(x_1, x_2) = \begin{array}{l} a(x_2 - bx_1^2 + cx_1 - d)^2 \\ + e(1-f)Cos(x_1) + e \end{array} \quad (17)$$

where $a = 1$, $b = \frac{5.1}{4\pi^2}$, $c = \frac{5}{\pi}$, $d = 6$, $e = 10$, and $f = \frac{1}{8\pi}$.

*13) Easom's function:* The optimal value for Easom's function is at $(x_1, x_2) = (\pi, \pi)$ and the optimum value is $-1$. The search space is usually restricted to $-100 \leq x_1 \leq 100$, $-100 \leq x_2 \leq 100$.

$$f_{eas}(x_1, x_2) = -Cos(x_1)Cos(x_2)e^{-(x_1-\pi)^2-(x_2-\pi)^2} \quad (18)$$

*14) De Jong Fifth function:* The search space is usually restricted to $-65.536 \leq x_i \leq 65.536$.

$$f_{djf}(x_1, x_2) = \begin{array}{l} \{0.002 + \sum_{i=-2}^{2}\sum_{j=-2}^{2}[ \\ 5(i+2) + j + 3 + (x_1 - 16j)^6 \\ + (x_2 - 16i)^6]^{-1}\}^{-1} \end{array} \quad (19)$$

*15) Shubert's function:* The search space is usually restricted to $-5.12 \leq x_i \leq 5.12$.

$$f_{shu}(x_1, x_2) = \begin{array}{l} -\sum_{i=1}^{5}iCos((i+1)x_1+1) \\ \times \sum_{i=1}^{5}iCos((i+1)x_2+1) \end{array} \quad (20)$$

## B. Results

In section V-A, some benchmark functions are introduced. In this section, the results of CSO on those functions are compared with those of other methods.

For the CSO method, each experiment is run 100 times. In all the figures, the best-so-far, mean, and median results are illustrated. The results for other methods are selected from corresponding tables. The corresponding identification numbers for research papers and methods are listed in Tables I and II.

Since the compared results are collected from different research papers, and because of a different floating Point Precision in different environments, all the values below $10^{-12}$ are represented by $(< 10^{-12})$ in all the experimental results, and are shown at the same level as $10^{-12}$ in the graphs. Note that not all research papers perform the experiment on all benchmark functions, and so the compared methods may differ for benchmark functions with different dimensions. The results for $f_{sph}$ are shown in Figures 12 and 12, and reported in Table III

| Research [#ref] | RID |
|---|---|
| This research | 0 |
| Kiranyaz2011 [39] | 1 |
| Xi2008 [40] | 2 |
| Rashedi2009 [41] | 3 |
| Gao2012 [42] | 4 |
| Karaboga2009 [43] | 5 |
| Brest2008 [44] | 6 |

TABLE I
RESEARCH IDs. FOR THE SAKE OF SAVING THE PAPER SPACE, A RESEARCH ID (RID) IS ASSIGNED TO EACH WORK.

| Global Optimization Method [#ref] | MID |
|---|---|
| PSO [3] | pso |
| bPSO [39] | bpso |
| SAD-PSO-A2 [39] | spso2 |
| WQPSO [40] | wqpso |
| RGA [41] | rga |
| GSA [41] | gsa |
| GA [43] | ga |
| ABC [43] | abc |
| DE [45] | de |
| jDEdynNP-F [44] | jde |
| CSO | cso |

TABLE II
METHOD IDs. FOR THE SAKE OF SAVING THE PAPER SPACE, A METHOD ID (MID) IS ASSIGNED TO EACH METHOD.

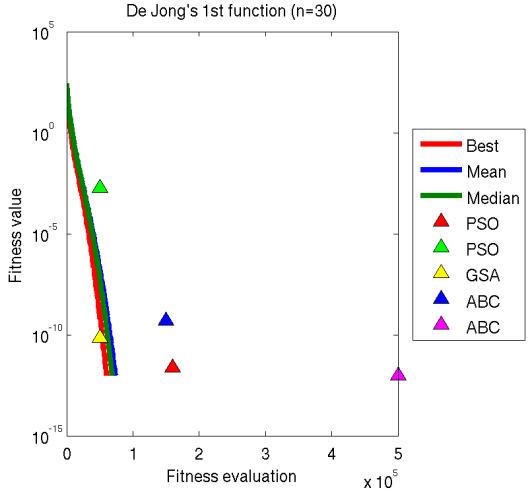

Fig. 12. CSO results for $f_{sph}$. ($n = 30$)

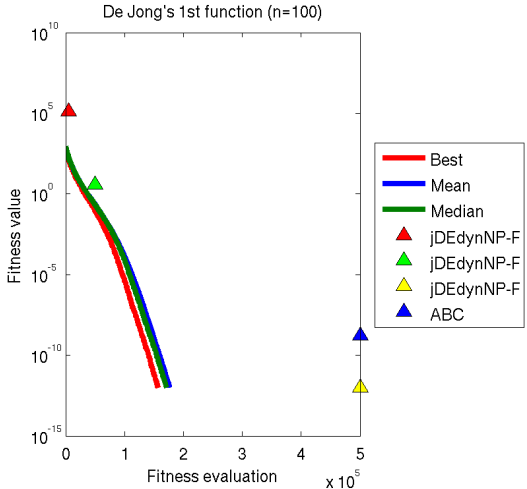

Fig. 13. CSO results for $f_{sph}$. ($n = 100$)

| RID | MID | Dim. | NFE / NFE (max) | Mean (St.D.) |
|---|---|---|---|---|
| 1 | pso | 20 | 400000 | $\leq$ 1e-12 ($\leq$ 1e-12) |
| 1 | spso2 | 20 | 400000 | $\leq$ 1e-12 ($\leq$ 1e-12) |
| 2 | wqpso | 20 | 120000 | 2.59e-74 (2.62e-76) |
| 2 | pso | 20 | 120000 | 2.68e-17 (5.24e-17) |
| 0 | cso | 20 | 63854 | $\leq$ 1e-12 ($\leq$ 1e-12) |
| 2 | wqpso | 30 | 160000 | 2.14e-60 (1.91e-62) |
| 2 | pso | 30 | 160000 | 2.47e-12 (7.16e-12) |
| 3 | rga | 30 | 50000 | 23.13 |
| 3 | pso | 30 | 50000 | 1.8e-3 |
| 3 | gsa | 30 | 50000 | 7.3e-11 |
| 4 | abc | 30 | 150000 | 5.21e-10 (2.46e-10) |
| 5 | ga | 30 | 500000 | 1.11e+3 (74.21) |
| 5 | pso | 30 | 500000 | $\leq$ 1e-12 ($\leq$ 1e-12) |
| 5 | de | 30 | 500000 | $\leq$ 1e-12 ($\leq$ 1e-12) |
| 5 | abc | 30 | 500000 | $\leq$ 1e-12 ($\leq$ 1e-12) |
| 0 | cso | 30 | 76101 | $\leq$ 1e-12 ($\leq$ 1e-12) |
| 1 | pso | 50 | 400000 | $\leq$ 1e-12 ($\leq$ 1e-12) |
| 1 | spso2 | 50 | 400000 | $\leq$ 1e-12 ($\leq$ 1e-12) |
| 0 | cso | 50 | 104160 | $\leq$ 1e-12 ($\leq$ 1e-12) |
| 6 | jde | 100 | 5000 | 136300 (13923) |
| 6 | jde | 100 | 50000 | 3.7501 (1.21) |
| 6 | jde | 100 | 500000 | 5.7e-14 (5.68e-14) |
| 4 | abc | 100 | 500000 | 1.64e-9 (9.85e-10) |
| 0 | cso | 100 | | $\leq$ 1e-12 ($\leq$ 1e-12) |

TABLE III
SPHERE FUNCTION $f_{sph}$ BENCHMARKS

The CSO method results for $f_{aph}$ with $n=30$ and $n=100$ are reflected in Table IV and shown in Figures 14 and 15. In Figure 15, which is similar to Figure 13, the CSO algorithm is slower at the start in terms of convergence, and has a higher speed after a turning point. These turning points are related to the Ant Colony-based width selection mechanism. This mechanism automatically focuses on exploration in the beginning and exploitation when there is no improvement in exploration.

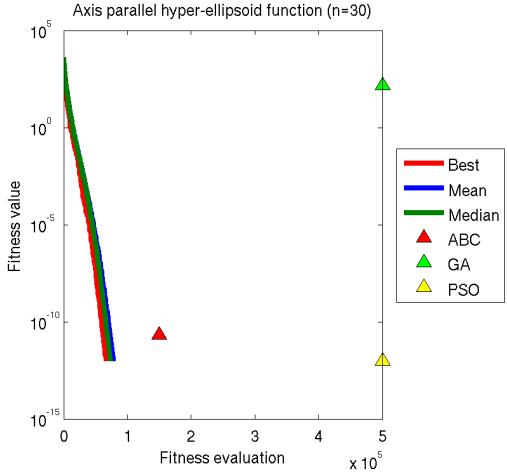

Fig. 14. CSO results for $f_{aph}$. ($n=30$)

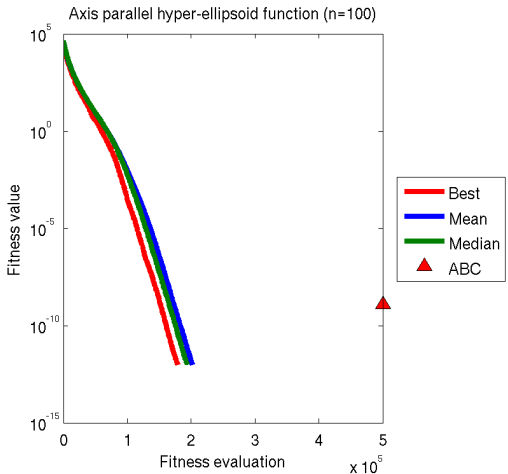

Fig. 15. CSO results for $f_{aph}$. ($n=100$)

| RID | MID | Dim. | NFE / NFE (max) | Mean (St.D.) |
|---|---|---|---|---|
| 4 | abc | 30 | 150000 | 2.22e-11 (1.14e-11) |
| 5 | ga | 30 | 500000 | 1.48e+2 (12.40) |
| 5 | pso | 30 | 500000 | $\leq$ 1e-12 ($\leq$ 1e-12) |
| 5 | de | 30 | 500000 | $\leq$ 1e-12 ($\leq$ 1e-12) |
| 5 | abc | 30 | 500000 | $\leq$ 1e-12 ($\leq$ 1e-12) |
| 0 | cso | 30 | 84603 | $\leq$ 1e-12 ($\leq$ 1e-12) |
| 4 | abc | 100 | 500000 | 1.25e-9 (9.75e-10) |
| 0 | cso | 100 | 206978 | $\leq$ 1e-12 ($\leq$ 1e-12) |

TABLE IV
AXIS PARALLEL HYPER-ELLIPSOID FUNCTION $f_{aph}$ BENCHMARKS

| RID | MID | Dim. | NFE / NFE (max) | Mean (St.D.) |
|---|---|---|---|---|
| 1 | bpso | 20 | 400000 | 1.264 (0.4382) |
| 1 | spso2 | 20 | 400000 | 1.299 (0.4658) |
| 2 | wqpso | 20 | 120000 | 47.02 (0.35) |
| 2 | pso | 20 | 120000 | 83.69 (137.2) |
| 0 | cso | 20 | 200000 | 6.4748 (4.9733) |
| 2 | wqpso | 30 | 160000 | 51.82 (0.31) |
| 2 | pso | 30 | 160000 | 202.6 (289.9) |
| 3 | rga | 30 | 50000 | 1.1e+3 |
| 3 | pso | 30 | 50000 | 3.6e+4 |
| 3 | gsa | 30 | 50000 | 25.16 |
| 4 | abc | 30 | 150000 | 4.23e-1 (4.34e-1) |
| 5 | ga | 30 | 500000 | 1.96e+5 (3.85e+4) |
| 5 | pso | 30 | 500000 | 15.08 (24.17) |
| 5 | de | 30 | 500000 | 18.20 (5.036) |
| 5 | abc | 30 | 500000 | 8.877e-2 (7.739e-2) |
| 0 | cso | 30 | 200000 | 28.5611 (25.9173) |
| 1 | bpso | 50 | 400000 | 15.90 (5.214) |
| 1 | spso2 | 50 | 400000 | 12.35 (2.677) |
| 0 | cso | 50 | 400000 | 52.3378 (34.255) |
| 6 | jde | 100 | 5000 | 4.562e+10 (8.589e+9) |
| 6 | jde | 100 | 50000 | 3.404e+4 (1.751e+4) |
| 6 | jde | 100 | 500000 | 1.115e+2 (4.476e+1) |
| 4 | abc | 100 | 500000 | 1.59 (1.23) |
| 0 | cso | 100 | 500000 | 152.08 (40.004) |

TABLE V
ROSENBROCK'S FUNCTION $f_{ros}$ BENCHMARKS

The CSO method results for $f_{ros}$ with $n=30$ and $n=100$ are reflected in Table V and shown in Figures 16 and 17. The only method that performs better than the CSO algorithm in this benchmark function is the ABC algorithm. However, the best results of CSO are better than ABC algorithm mean results, but needs to be compared with ABC best results, if known.

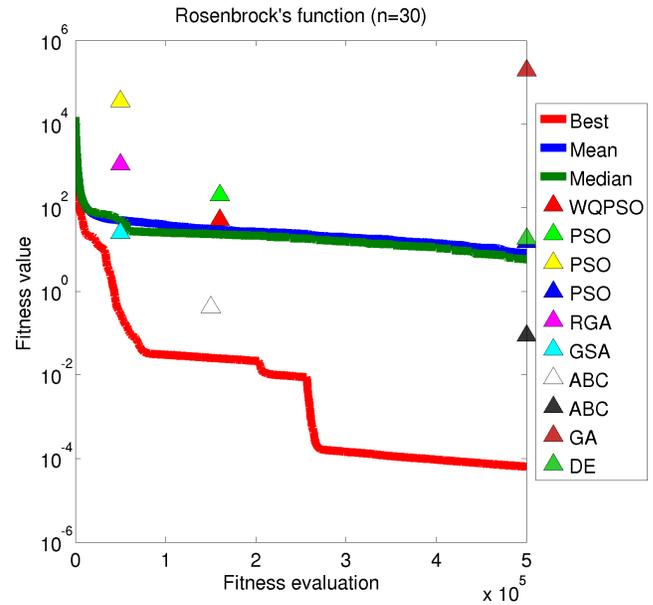

Fig. 16. CSO results for $f_{ros}$. ($n=30$)

| RID | MID | Dim. | NFE / NFE (max) | Mean (St.D.) |
|---|---|---|---|---|
| 1 | pso | 20 | 400000 | 4.29e-2 (3.83e-2) |
| 1 | spso2 | 20 | 400000 | 3.83e-2 (3.69e-2) |
| 2 | wqpso | 20 | 120000 | 7.28 (0.0032) |
| 2 | pso | 20 | 120000 | 13.38 (8.51) |
| 0 | cso | 20 | 120780 | $\leq$ 1e-12 ($\leq$ 1e-12) |
| 2 | wqpso | 30 | 160000 | 15.02 (0.0294) |
| 2 | pso | 30 | 160000 | 28.62 (10.34) |
| 3 | rga | 30 | 50000 | 5.9 |
| 3 | pso | 30 | 50000 | 55.1 |
| 3 | gsa | 30 | 50000 | 15.32 |
| 4 | abc | 30 | 150000 | 4.81e-3 (2.57e-2) |
| 5 | ga | 30 | 500000 | 52.92 (4.56) |
| 5 | pso | 30 | 500000 | 43.97 (11.72) |
| 5 | de | 30 | 500000 | 11.71 (2.53) |
| 5 | abc | 30 | 500000 | $\leq$ 1e-12 ($\leq$ 1e-12) |
| 0 | cso | 30 | 197922 | $\leq$ 1e-12 ($\leq$ 1e-12) |
| 1 | pso | 50 | 400000 | 5.28e-2 (6.88e-2) |
| 1 | spso2 | 50 | 400000 | 3.81e-2 (4.36e-2) |
| 0 | cso | 50 | 273847 | $\leq$ 1e-12 ($\leq$ 1e-12) |
| 6 | jde | 100 | 5000 | 1.288e+3 (4.560e+1) |
| 6 | jde | 100 | 50000 | 3.425e+2 (2.454e+1) |
| 6 | jde | 100 | 500000 | 5.457e-14 (1.136e-14) |
| 4 | abc | 100 | 500000 | 1.1 (8.21e-1) |
| 0 | cso | 100 | 500000 | 0.0610 (0.2807) |

TABLE VI
RASTRIGIN'S FUNCTION $f_{ras}$ BENCHMARKS

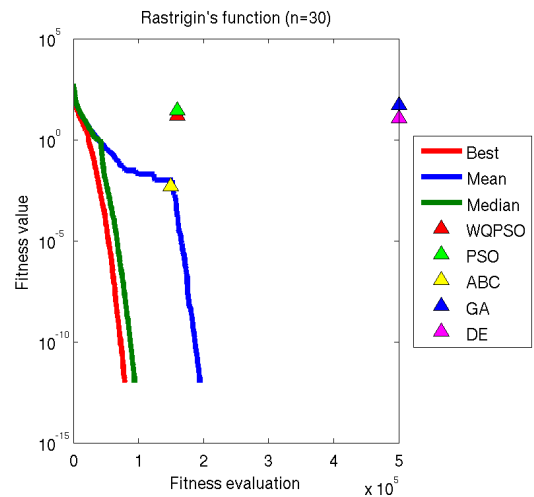

Fig. 18. CSO results for $f_{ras}$. ($n = 30$)

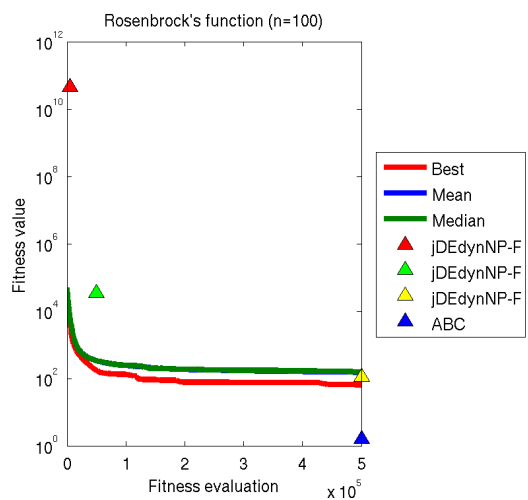

Fig. 17. CSO results for $f_{ros}$. ($n = 100$)

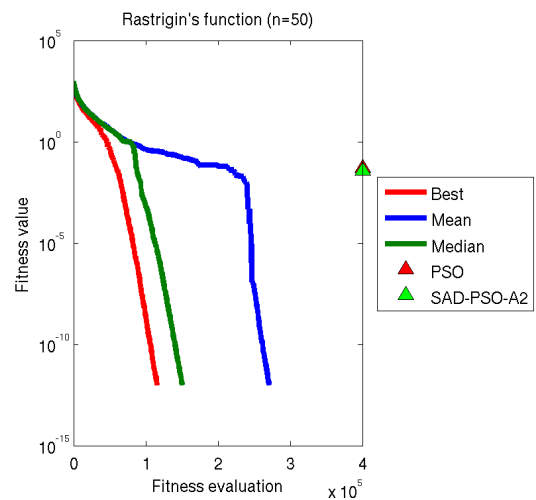

Fig. 19. CSO results for $f_{ras}$. ($n = 50$)

As shown in Figure 18 and Figure 19, the CSO algorithm results are better than those of traditional algorithms such as GA and DE. The results of the ABC algorithm are comparable to the CSO results. The graph of median values is near the best results, which shows that more than 50% of the results perform near the best CSO results. In Rosenbrocks function, the graph of median values is near the graph of mean values, which indicates the reverse, as shown in Figure 16. The CSO method results for $f_{sch}$ with $n = 30$ and $n = 100$ are reflected in Table VII and are shown in Figures 20 and 21.

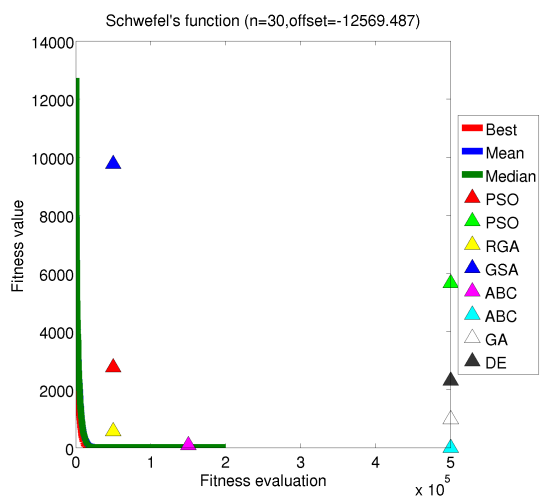

Fig. 20. CSO results for $f_{sch}$. ($n = 30$)

| RID | MID | Dim. | NFE / NFE (max) | Mean (St.D.) |
|---|---|---|---|---|
| 1 | pso | 20 | 400000 | -8377.9106 (0.3915) |
| 1 | spso2 | 20 | 400000 | -8379.3504 (0.0758) |
| 0 | cso | 20 | 200000 | -8379.657745 (7.697e-12) |
| 3 | rga | 30 | 50000 | -12000 |
| 3 | pso | 30 | 50000 | -9800 |
| 3 | gsa | 30 | 50000 | -2800 |
| 4 | abc | 30 | 150000 | -12480.887 (8.62e+1) |
| 5 | ga | 30 | 500000 | -11593.4 (93.25) |
| 5 | pso | 30 | 500000 | -6909.13 (457.9) |
| 5 | de | 30 | 500000 | -10266 (521.84) |
| 5 | abc | 30 | 500000 | -12569.487 ($\leq$ 1e-12) |
| 0 | cso | 30 | 200000 | -12569.486618 (1.1e-11) |
| 1 | pso | 50 | 400000 | -20938.9423 (2.2145) |
| 1 | spso2 | 50 | 400000 | -20948.3172 (0.1093) |
| 0 | cso | 50 | 400000 | -20949.1443 (3.8-11) |
| 6 | jde | 100 | 5000 | -41777.09 (3.736) |
| 6 | jde | 100 | 50000 | -41861.37 (3.265) |
| 6 | jde | 100 | 500000 | -41897.8613 (2.270e-1) |
| 4 | abc | 100 | 500000 | -40608.29 (2.23e+2) |
| 0 | cso | 100 | 500000 | -41898.28872 (6.9-11) |

TABLE VII
SCHWEFEL'S $f_{sch}$ FUNCTION

| RID | MID | Dim. | NFE / NFE (max) | Mean (St.D.) |
|---|---|---|---|---|
| 1 | pso | 20 | 400000 | $\leq$ 1e-12 ($\leq$ 1e-12) |
| 1 | spso2 | 20 | 400000 | $\leq$ 1e-12 ($\leq$ 1e-12) |
| 2 | wqpso | 20 | 120000 | 3.25e-4 (4.17e-4) |
| 2 | pso | 20 | 120000 | 0.02854 (0.0268) |
| 0 | cso | 20 | 400000 | 0.02144 (0.02538) |
| 2 | wqpso | 30 | 160000 | 4.22e-5 (1.37e-5) |
| 2 | pso | 30 | 160000 | 0.01258 (0.01396) |
| 3 | rga | 30 | 50000 | 1.16 |
| 3 | pso | 30 | 50000 | 0.01 |
| 3 | gsa | 30 | 50000 | 0.29 |
| 4 | abc | 30 | 150000 | 1.61e-8 (3.99e-8) |
| 5 | ga | 30 | 500000 | 10.63 (1.161) |
| 5 | pso | 30 | 500000 | 1.739e-2 (2.080e-2) |
| 5 | de | 30 | 500000 | 1.479e-3 (2.958e-3) |
| 5 | abc | 30 | 500000 | $\leq$ 1e-12 ($\leq$ 1e-12) |
| 0 | cso | 30 | 200000 | 0.01448 ( 0.01623) |
| 1 | pso | 50 | 400000 | 50.73 (191.1) |
| 1 | spso2 | 50 | 400000 | $\leq$ 1e-12 ($\leq$ 1e-12) |
| 0 | cso | 50 | 400000 | 0.007855 (0.01017) |
| 6 | jde | 100 | 5000 | 1.116e+3 (9.243e+1) |
| 6 | jde | 100 | 50000 | 9.087e-1 (6.561e-2) |
| 6 | jde | 100 | 500000 | 2.842e-14 ($\leq$ 1e-12) |
| 4 | abc | 100 | 500000 | 2.01e-9 (1.32e-9) |
| 0 | cso | 100 | 500000 | 0.005146 ( 0.008913) |

TABLE VIII
GRIEWANGK'S FUNCTION $f_{gri}$ BENCHMARKS

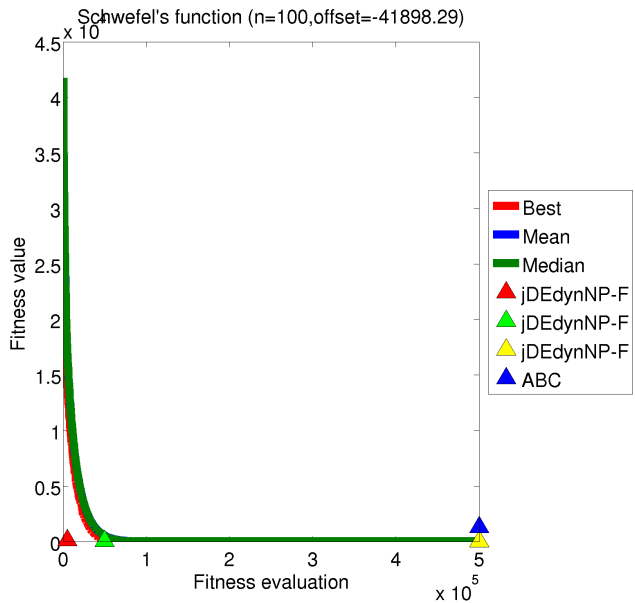

Fig. 21. CSO results for $f_{sch}$. ($n = 100$)

The CSO method results for $f_{gri}$ with $n = 30$ and $n = 100$ are presented in Table VIII and shown in Figures 22 and 23, and the results of the CSO method for $f_{sdp}$ and $f_{ack}$ are reflected in Table IX and Table X, and shown in Figures 24, 25, 26, and 27 respectively. The CSO performs better in all cases. The results of the GSA method in Ackleys function ($n = 30$) are comparable to the CSO algorithm results.

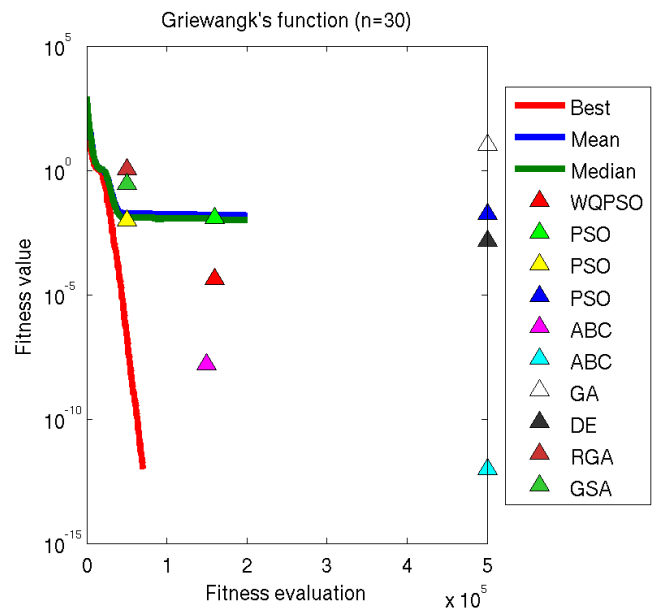

Fig. 22. CSO results for $f_{gri}$. ($n = 30$)

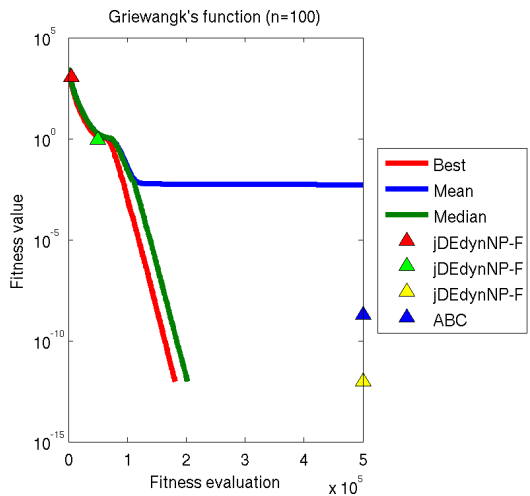

Fig. 23. CSO results for $f_{gri}$. ($n = 100$)

| RID | MID | Dim. | NFE / NFE (max) | Mean (St.D.) |
|---|---|---|---|---|
| 4 | abc | 30 | 150000 | 1.45e-16 (1.55e-16) |
| 0 | cso | 30 | 55855 | $\leq$ 1e-12 ($\leq$ 1e-12) |
| 4 | abc | 100 | 500000 | 4.83e-7 (7.88e-7) |
| 0 | cso | 100 | 142893 | $\leq$ 1e-12 ($\leq$ 1e-12) |

TABLE IX
SUM OF DIFFERENT POWER FUNCTION $f_{sdp}$ BENCHMARKS

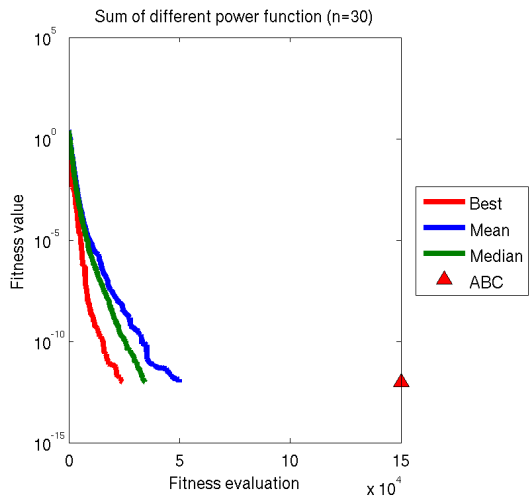

Fig. 24. CSO results for $f_{sdp}$. ($n = 30$)

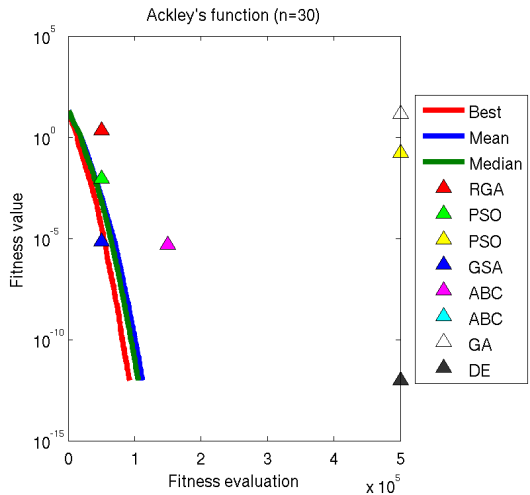

Fig. 26. CSO results for $f_{ack}$. ($n = 30$)

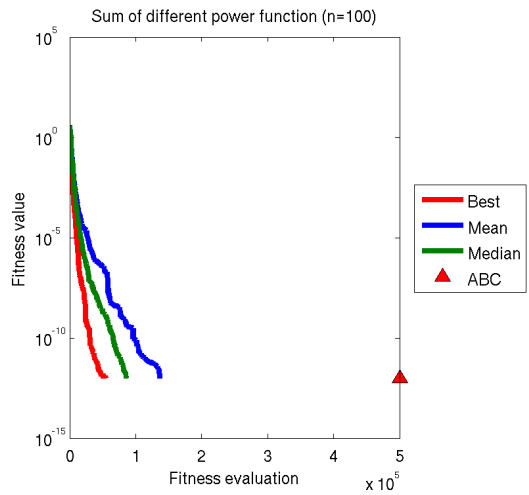

Fig. 25. CSO results for $f_{sdp}$. ($n = 100$)

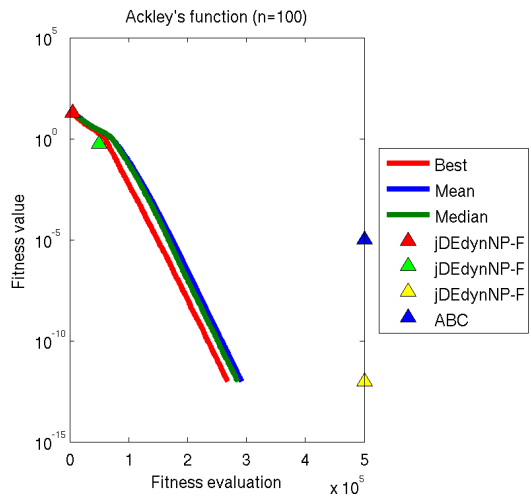

Fig. 27. CSO results for $f_{ack}$. ($n = 100$)

| RID | MID | Dim. | NFE / NFE (max) | Mean (St.D.) |
|---|---|---|---|---|
| 3 | rga | 30 | 50000 | 2.13 |
| 3 | pso | 30 | 50000 | 9.0e-3 |
| 3 | gsa | 30 | 50000 | 6.9e-6 |
| 4 | abc | 30 | 150000 | 4.83e-6 (2.12e-6) |
| 5 | ga | 30 | 500000 | 14.67 (1.781e-1) |
| 5 | pso | 30 | 500000 | 1.646e-1 (4.938e-1) |
| 5 | de | 30 | 500000 | $\leq$ 1e-12 ($\leq$ 1e-12) |
| 5 | abc | 30 | 500000 | $\leq$ 1e-12 ($\leq$ 1e-12) |
| 0 | cso | 30 | 114992 | $\leq$ 1e-12 ($\leq$ 1e-12) |
| 6 | jde | 100 | 5000 | 2.013e+1 (1.621e-1) |
| 6 | jde | 100 | 50000 | 5.872e-1 (1.051e-1) |
| 6 | jde | 100 | 500000 | $\leq$ 1e-12 ($\leq$ 1e-12) |
| 4 | abc | 100 | 500000 | 1.02e-5 (2.92e-6) |
| 0 | cso | 100 | 296187 | $\leq$ 1e-12 ($\leq$ 1e-12) |

TABLE X
ACKLEY'S FUNCTION $f_{ack}$ BENCHMARKS

| RID | MID | Dim. | NFE / NFE (max) | Mean (St.D.) |
|---|---|---|---|---|
| 3 | rga | 30 | 50000 | 5.6e+3 |
| 3 | pso | 30 | 50000 | 4.1e+3 |
| 3 | gsa | 30 | 50000 | 0.16e+3 |
| 0 | cso | 30 | 50000 | 3.789e-4 (5.525e-4) |

TABLE XI
ROTATED HYPER-ELLIPSOID FUNCTION $f_{rhe}$ BENCHMARKS

The CSO method results for $f_{rhe}$ with $n = 30$ are reflected in Table XI and shown in Figure 28, revealing that CSO performs better than the RGA, PSO, and GSA algorithms. The CSO results for $f_{mic}$, $f_{bra}$, $f_{eas}$, $f_{djf}$, and $f_{shu}$ are reflected in Table XII, Table XIV, Table XV, Table XVI, and Table XVII respectively, and shown in Figure 29, Figure 30, Figure 31, Figure 32, and Figure 33.

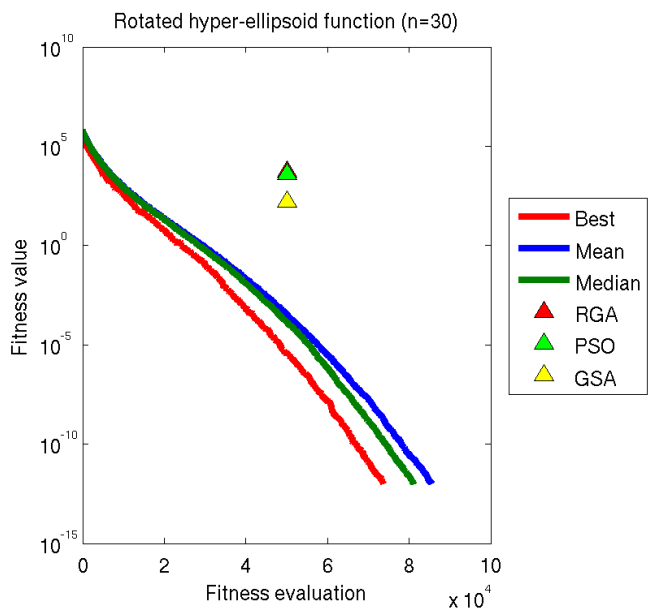

Fig. 28. CSO results for $f_{rhe}$. ($n = 30$)

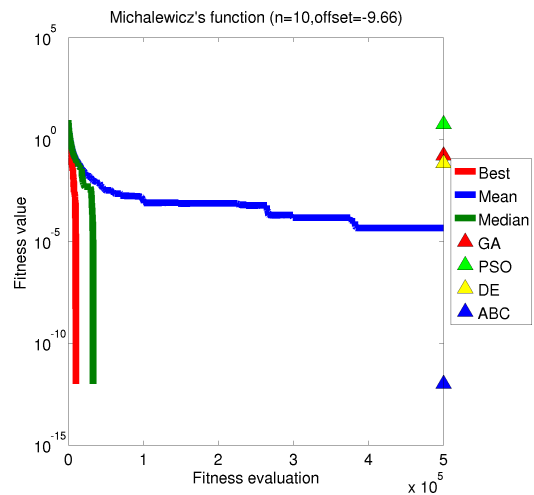

Fig. 29. CSO results for $f_{mic}$. ($n = 10$)

| RID | MID | Dim. | NFE / NFE (max) | Mean (St.D.) |
|---|---|---|---|---|
| 5 | ga | 10 | 500000 | -9.49683 (0.1411) |
| 5 | pso | 10 | 500000 | -4.0071803 (0.5026) |
| 5 | de | 10 | 500000 | -9.591151 (0.06420) |
| 5 | abc | 10 | 500000 | -9.6601517 ($\leq$ 1e-12) |
| 0 | cso | 10 | 100000 | -9.6588 (0.005780) |

TABLE XII
MICHALEWICZ'S FUNCTION $f_{mic}$ BENCHMARKS

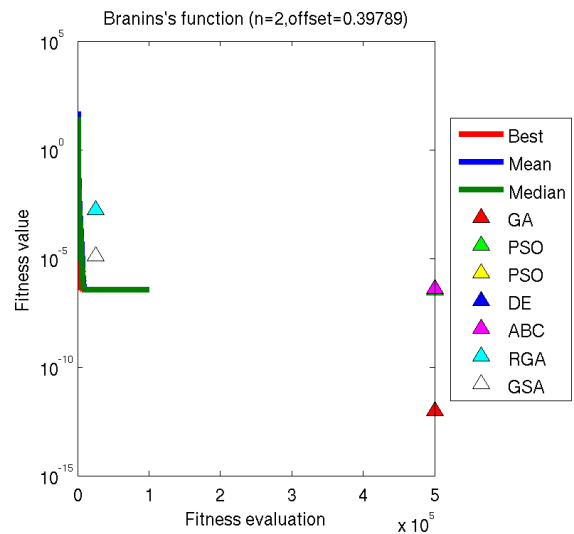

Fig. 30. CSO results for $f_{bra}$. ($n = 2$)

| RID | MID | Dim. | NFE / NFE (max) | Mean (St.D.) |
|---|---|---|---|---|
| 0 | cso | 30 | 38569 | -78.3323 (2.77e-5) |
| 5 | abc | 100 | 150000 | -77.5964 (2.23e-1) |
| 0 | cso | 100 | 104844 | -78.3323 (3.61e-5) |

TABLE XIII
2N-MINIMA FUNCTION $f_{2nm}$ BENCHMARKS

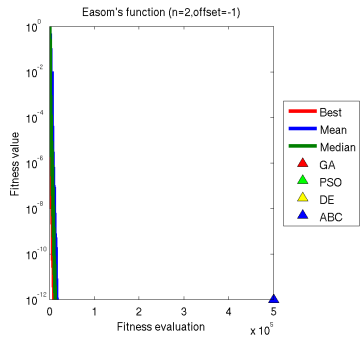

Fig. 31. CSO results for $f_{eas}$. ($n = 2$)

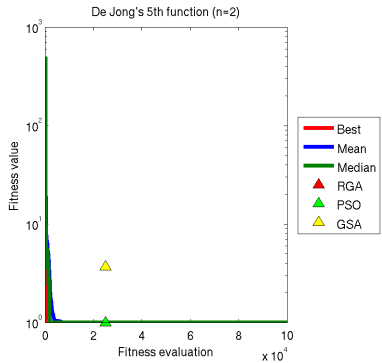

Fig. 32. CSO results for $f_{djf}$. ($n = 2$)

For two-dimensional benchmark functions, the CSO algorithm performs very well and convergence is very rapid compared to other methods. In Figure 30, the CSO algorithm reaches values below $10^{-12}$, which are not shown on the graph.

| RID | MID | Dim. | NFE / NFE (max) | Mean (St.D.) |
|---|---|---|---|---|
| 5 | ga | 2 | 500000 | 0.397887 ($\leq$ 1e-12) |
| 5 | pso | 2 | 500000 | 0.39788736 ($\leq$ 1e-12) |
| 5 | de | 2 | 500000 | 0.3978874 ($\leq$ 1e-12) |
| 5 | abc | 2 | 500000 | 0.3978874 ($\leq$ 1e-12) |
| 3 | rga | 2 | 25000 | 0.3996 |
| 3 | pso | 2 | 25000 | 0.3979 |
| 3 | gsa | 2 | 25000 | 0.3979 |
| 0 | cso | 2 | 15000 | 0.3978873577 ($\leq$ 1e-12) |

TABLE XIV
BRANINS'S FUNCTION $f_{bra}$ BENCHMARKS

| RID | MID | Dim. | NFE / NFE (max) | Mean (St.D.) |
|---|---|---|---|---|
| 5 | ga | 2 | 500000 | -1 ($\leq$ 1e-12) |
| 5 | pso | 2 | 500000 | -1 ($\leq$ 1e-12) |
| 5 | de | 2 | 500000 | -1 ($\leq$ 1e-12) |
| 5 | abc | 2 | 500000 | -1 ($\leq$ 1e-12) |
| 0 | cso | 2 | 22967 | -1 ($\leq$ 1e-12) |

TABLE XV
EASOM'S FUNCTION $f_{eas}$ BENCHMARKS

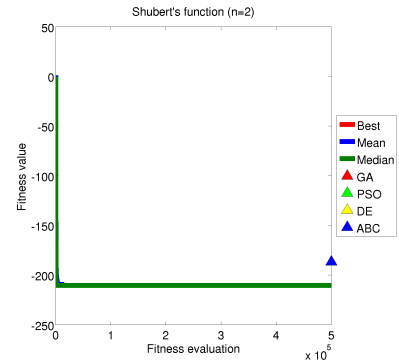

Fig. 33. CSO results for $f_{shu}$. ($n = 2$)

| RID | MID | Dim. | NFE / NFE (max) | Mean (St.D.) |
|---|---|---|---|---|
| 3 | rga | 2 | 25000 | 0.998 |
| 3 | pso | 2 | 25000 | 0.998 |
| 3 | gsa | 2 | 25000 | 3.70 |
| 0 | cso | 2 | 10000 | 0.99800 (1.7-10) |

TABLE XVI
DE JONG FIFTH FUNCTION $f_{djf}$ BENCHMARKS

## VI. CONCLUSION AND FUTURE WORK

In this work, a novel global optimization method called Curved Space Optimization (CSO) has been introduced, and a simple implementation of this method has been tested on various state-of-the-art benchmark functions. The method is based on Random Search with some transformations in its search space, based on general relativity theory. The search space transformations are defined as curvatures around previously searched spots. These curvatures will lead the method towards a global optimum, which makes Random Search an effective method with exploitation capability. Several adaptive mechanisms are used to control the depth and radius of curvatures during the search process.

As shown in the experimental results section, CSO performs better on most of the benchmark functions than well known state-of-the-art methods. For some benchmark functions, such as Michalewiczs and Griewangks (n=100) functions, CSOs best and median results are a great deal better than its mean results, which indicates that CSO is over 50% more effective in these functions, however the CSO mean results are also better than most of the benchmark functions that were compared. The algorithm performs well on both low- and high-dimensional functions, and on both unimodal and multimodal functions. The exploration-exploitation tradeoff in the CSO algorithm can be seen in most results graphs as a turning point, where the graph curvature changes. This tradeoff is controlled by an adaptive mechanism based on the Ant Colony method to maximize the performance of the algorithm.

| RID | MID | Dim. | NFE / NFE (max) | Mean (St.D.) |
|---|---|---|---|---|
| 5 | ga | 2 | 500000 | -186.731 ($\leq$ 1e-12) |
| 5 | pso | 2 | 500000 | -186.73091 ($\leq$ 1e-12) |
| 5 | de | 2 | 500000 | -186.7309 ($\leq$ 1e-12) |
| 5 | abc | 2 | 500000 | -186.73091 ($\leq$ 1e-12) |
| 0 | cso | 2 | 100000 | -210.4822 ($\leq$ 1e-12) |

TABLE XVII
SHUBERT'S FUNCTION $f_{shu}$ BENCHMARKS

As a prospect for future research, the effect of various implementations of the CSO method on its performance will be studied. This will include a study of the variety of curvature shapes and a study of the various mechanisms for selecting the depth and width of curvatures.

ACKNOWLEDGMENT

The authors thank CANARIE (Canadian Network for Advanced Research in Education) for their financial support.